\documentclass{article}
\usepackage{ijcai22}

\usepackage{bm}
\usepackage{siunitx}
\usepackage{times}
\usepackage{soul}
\usepackage{url}
\usepackage[hidelinks]{hyperref}
\usepackage[utf8]{inputenc}
\usepackage[small]{caption}
\usepackage{graphicx}
\usepackage{amsmath}
\usepackage{amsthm}
\usepackage{booktabs}
\usepackage{algorithm}
\usepackage{algorithmic}
\usepackage{wrapfig}
\usepackage{cleveref}       
\usepackage{natbib}
\usepackage{subfig}
\usepackage{amssymb}

\title{Behavioral experiments for understanding catastrophic forgetting}

\author{
Samuel J.\ Bell
\And
Neil D.~Lawrence \\
\affiliations
Computer Laboratory, University of Cambridge \\
\emails
\texttt{\{sjb326,ndl21\}@cam.ac.uk}
}

\graphicspath{{fig/}}

\begin{document}

\maketitle


\begin{abstract}
    We explore how the fundamental tool of experimental psychology, the behavioral experiment, can generate insight not only into humans and animals, but artificial systems too.
    Using catastrophic forgetting as a case study, we present a series of novel experiments with two-layer non-linear networks, and exploratory results revealing a more nuanced understanding of this complex behavior.
    Alongside our empirical findings, we demonstrate and discuss an alternative, behavior-first approach to investigating emergent neural network phenomena.
\end{abstract}


\section{Introduction}

Neural networks have evolved into a staple of contemporary machine learning.
Yet, as with any complex system, they can be challenging to interpret and understand, often exhibiting surprising emergent behavior that defies current theoretical accounts.

Humans and animals are also complex systems, and since the mid-nineteenth century the science of experimental psychology has developed methods for investigating their behavior.
From the limits of perception \citep{Peirce1885}, to the capacity of short-term memory \citep{Miller1956}, the fundamental tool of the experimental psychologist is the behavioral experiment.
In this work, we demonstrate the value of the behavioral experiment---and more broadly, ``thinking like a psychologist''---to the machine learning community through a case study into a specific neural network phenomenon.

Catastrophic forgetting (CF) is where training on a new task degrades performance on previously learned tasks \citep{McCloskey1989, Ratcliff1990, French1999, Parisi2019, Hadsell2020}.
The often dramatic decline in previous task performance, while the subject of numerous studies, is still only partially understood both empirically and theoretically.
For example, community consensus is absent around such core questions as: ``Does CF always occur?'' ``What causes it?'' and ``Can we predict its emergence?''

Using a two-layer ReLU network as a model organism, we devise a series of experiments with synthetic visual classification tasks.
There are a growing number of empirical CF investigations (see \cref{sec:background-empirical}), though almost all rely on modifications to pre-existing, naturalistic datasets such as MNIST \citep{LeCun2010}, CIFAR-10/100 \citep{Krizhevsky2009}, or ImageNet \citep{Russakovsky2015}.
In contrast, we precisely control both surface form and underlying distribution for each of our tasks, and are therefore able to manipulate the exact dimensions of task similarity---an impossible feat when using an off-the-shelf dataset.
This distinction, not between naturalistic and synthetic but between \emph{observation} and \emph{intervention}, is the crux of the utility of the behavioral experiment to machine learning.
Without intervention, we can only make claims about association, not cause.

Our experiments are a novel approach to a well-studied phenomenon, and our findings add nuance to a complex picture.
Our results indicate that CF is situation-specific; reveal the differential effects of perceptual and semantic task similarity; and demonstrate that the new task's loss surface is associated with forgetting.
Equipped with a sample of trained models, we investigate how their parameters have changed through time, and as a result identify a simple heuristic based on the gradients of the new task's loss function---after just a single backward pass---that is strongly predictive of CF\@.
We validate our findings on a more naturalistic dataset, a randomized version of split MNIST.
Finally, we prove that our claims generalize to a more complex model, ResNet-18 \citep{He2016}, trained on a randomized version of split CIFAR-10.

\section{Related work}

\subsection{Connectionism}

There is a long tradition of connectionist investigations into CF \citep{McCloskey1989, Ratcliff1990, French1993, Robins1995, French1999}.
However, while we have an empirical focus in common, our aims fundamentally differ.
Rather than investigating neural networks in search of psychological plausibility, here we use the \emph{methods} of psychology to investigate neural networks.
We make no claim as to the status of the neural network as model of mind or brain, we only suggest that the behavior-first approach is a useful experimental paradigm.

Connectionists probing CF (often using two-layer neural networks) have investigated ideas including representational overlap \citep{French1993}; the importance of overwritten parameters \citep{Sutton1986}; and previous task rehearsal \citep{Robins1995}.
These investigations involved small amounts of low-dimensional and noiseless data, which is an unrealistic setting for contemporary practice.
Research of this period is also likely to rely upon online gradient descent, rather than the stochastic mini-batch variant used in practice today.

\subsection{Empirical investigations}
\label{sec:background-empirical}

We are by no means the first to conduct empirical research into CF\@.
More recent investigation has involved analysis of the effect of hyperparameters \citep{Goodfellow2014}; mitigation strategies \citep{Kemker2018}; and benchmark datasets \citep{Zenke2017, Ramasesh2020}.
Unlike previous work, however, we treat the neural network as a holistic system, explicitly designing tasks and experiments to understand its behavior.
In contrast, experiments with split/permuted MNIST/CIFAR tell us little about the circumstances under which CF may take place, given the difficulty of precisely quantifying similarity between tasks.\footnote{For example, is a 0 vs.\ 1 digit classification more or less similar to 2 vs.\ 3 than to 4 vs.\ 5? In which ways are these tasks similar, and how do they differ?}
\citet{Ramasesh2020} investigate learned representations across tasks, distinguishing between surface form and semantic similarity, and suggesting evidence that moving between semantically-similar tasks leads to less CF.
We follow this line of investigation into semantic and perceptual differences, though use synthetic data to precisely control task similarity.

\citet{Lee2021} outline both empirical experiments in a teacher-student setup and an analytical account of continual learning dynamics.
The authors disentangle ``feature-level'' from ``readout-level'' task similarity (i.e.\ perceptual from semantic), defined as the teacher-student overlap in the input (``feature'') and output (``readout'') layer weights.
Our experiments are designed to take these distinct notions of similarity and translate them into new datasets of noisy, high-dimensional images that more closely approximate a real deep learning setting.

\subsection{Mitigation attempts}

While our work focuses on understanding, a significant body of research trials engineering approaches for overcoming CF.
Techniques employed fall broadly into regularization strategies \citep{French1993, Kirkpatrick2017, Zenke2017, Li2018}; capacity allocation \citep{Goodrich2014}, and rehearsal or replay \citep{Robins1995, French2002}.
As our study investigates the behavioral phenomenon of CF, we do not present a detailed survey of such work.
We do however highlight the similarity of our gradient-based heuristic to the regularizing term of \citet{Kirkpatrick2017}, whose \emph{Elastic Weight Consolidation} also makes use of the diagonal of the Fisher information matrix to approximate the Hessian of the loss surface.
\citet{Mirzadeh2020} also investigate the curvature of the first task loss, testing hyperparameter combinations that should lead to minima with wider curvature, and hence mitigate CF when tested on split/rotated MNIST and CIFAR-10\@.
Our work differs in its focus on the new task loss surface, rather than the former (see \cref{sec:methods-grads}).

\subsection{Behavioral testing}

Beyond continual learning, a broad spectrum of research applies purposely-designed experiments for black-box behavioral analysis.
Behavioral experiments have been used to investigate neural network generalization \citep{Neyshabur2017, Power2022}; memorization \citep{Zhang2017}; transfer learning \citep{Neyshabur2020}; adversarial robustness \citep{Szegedy2013}; and biased decision making \citep{Buolamwini2018} among many other topics.
In natural language processing, the behavioral experiment is already a critical tool for model analysis, such as experiments involving synthetic noise \citep{Belinkov2018}; stress testing \citep{Naik2018}; adversarial constructions \citep{Jia2017}; and purpose-built diagnostic datasets \citep{Johnson2017}.
From a practical perspective, \citet{Ribeiro2021} recently introduced a black-box framework, \emph{CheckList}, for systematically testing failure cases in NLP systems.
To re-emphasize an earlier point, these works do not search for ``human-like'' behaviors, but apply targeted behavioral experiments to answer pertinent questions in machine learning research and practice.
Here, we seek to do the same.


\section{Methods}

\begin{figure*}
    \centering
    \subfloat[Semantic shift]{
        \includegraphics[trim={0 1cm 0 1cm},clip,width=0.45\textwidth]{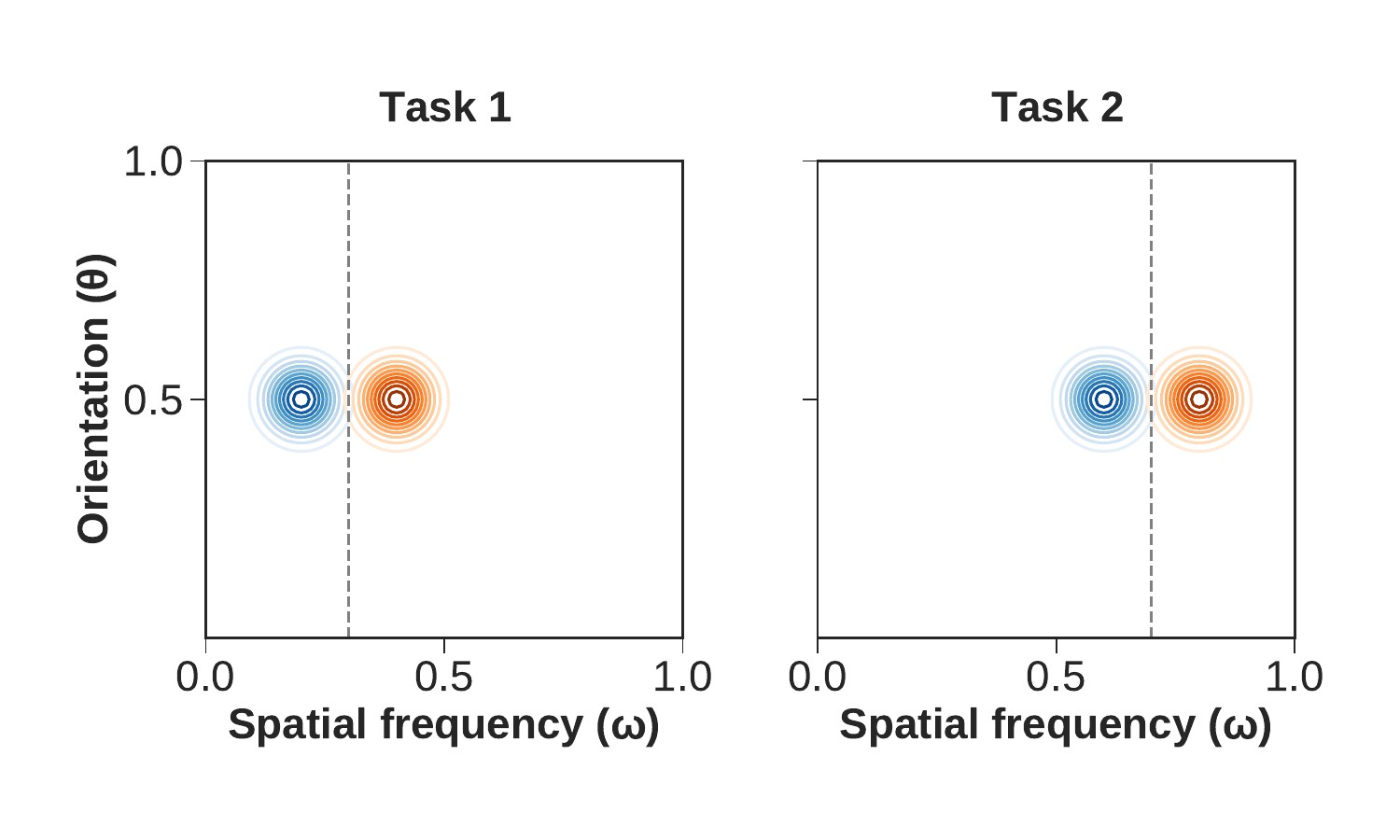}
    }\qquad
    \subfloat[Perceptual shift]{
        \includegraphics[trim={0 1cm 0 1cm},clip,width=0.45\textwidth]{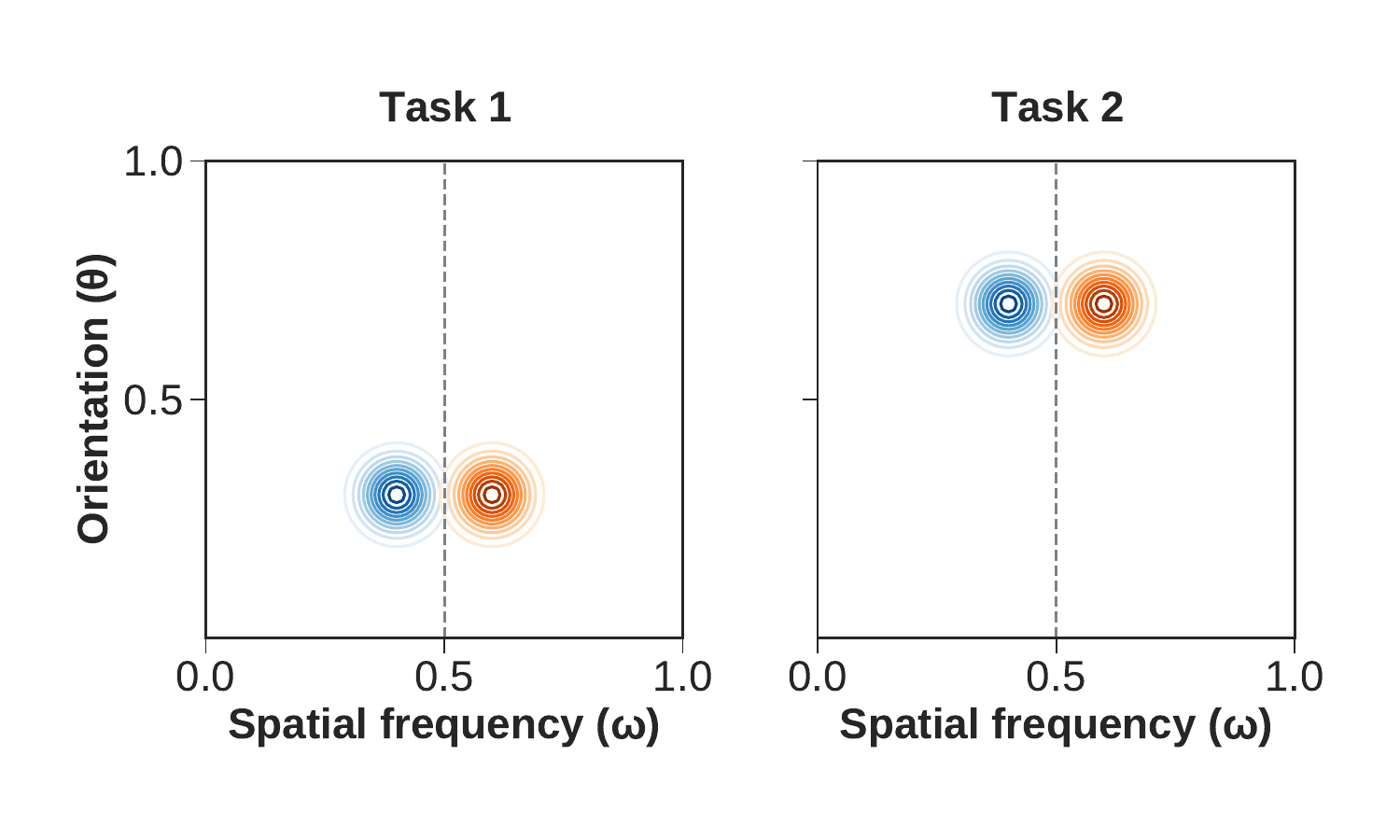}
    }
    \caption{Illustration of contrasting curricula of binary classification tasks (blue vs.\ orange). All categories are separable by $\omega$; decision boundary is dashed gray line. \textbf{(a)} Tasks in curriculum $\mathcal{C}_S$ differ semantically: the decision boundary shifts from task to task. \textbf{(b)} Tasks in $\mathcal{C}_P$ are semantically similar and share a decision boundary, yet differ perceptually due to varying \(\theta\). See \cref{sec:app-category-parameters} for actual distribution parameters. Though denoted as discrete tasks for simplicity, both curricula exemplify distribution shifts commonly found in practice.}\label{fig:curricula}
\end{figure*}

We now describe our five behavioral experiments designed to investigate CF\@.
Each experiment involves different types of ``curricula'': sequences of tasks that differ in specific ways.
After training two-layer ReLU networks on these curricula, we analyze the resulting network behavior by measuring validation losses.
Starting with our behavioral results, and a population of models to match, we then analyze parameter change throughout training on each task, and the associated loss function gradients.

Given two tasks $\mathcal{T}_j$ and $\mathcal{T}_{j+1}$, let $\mathcal{M}_{\bm{\theta}}$ be a parameterized model to be sequentially trained on both tasks.
We denote the model's parameters after training on task $j$ and $j+1$ as $\bm{\theta}_j$ and $\bm{\theta}_{j+1}$ respectively.
Assuming our model is optimized to minimize some loss function for each task $L_{j}(\bm{\theta}) \in \mathbb{R}$, we define CF as a function of the two tasks,
\begin{align*}
    F(\mathcal{T}_j, \mathcal{T}_{j+1}) = \begin{cases}
        1 & \quad \text{if } L_{j}(\bm{\theta}_{j+1}) - L_{j}(\bm{\theta}_{j}) > \epsilon \\
        0 & \quad \text{otherwise,}
    \end{cases}
\end{align*}
where $\epsilon$ is some application-specific acceptability threshold.
Thus, we test for CF by training a model on task $\mathcal{T}_j$ followed by $\mathcal{T}_{j+1}$.
Then, without updating its parameters, we re-evaluate performance of the trained model on the original task $\mathcal{T}_j$.
This sequential training paradigm forms the basis of all our experiments.

\subsection{Synthetic experiments: grating classification}

\citet{Kattner2017} investigate human transfer learning using sequences of classification tasks, where each task comprises linearly-separable categories defined by 2D Gaussians.
In our first three experiments, we follow this approach, defining a curriculum $\mathcal{C}$ as a set of tasks,
\begin{align*}
    \mathcal{C} = \{\: \mathcal{T}_j \mid 1 \leq j \leq N \:\} \ , 
\end{align*}
where each task $\mathcal{T}_{j}$ is a pair of 2D Gaussian distributions for its two categories,
\begin{align*}
    \mathcal{T}_{j} = \{\: \mathcal{N}(\bm{\mu}_{j}, \bm{\Sigma}), \ \mathcal{N}(\bm{\upsilon}_{j}, \bm{\Sigma}) \:\} \ .
\end{align*}
with each category parameterized by its mean vector $\bm{\mu}_j$ or $\bm{\upsilon}_j$ and covariance matrix $\bm{\Sigma} = I$.
We manipulate task similarity by modifying the difference between mean vectors.

\begin{figure}[b]
    \centering
    \includegraphics[trim={0 0 10cm 0 },clip,width=0.28\textwidth]{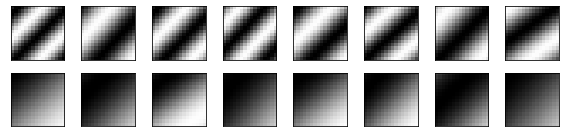}
    \caption{Image samples from a binary classification task over sinusoidal gratings. Rows are different categories.}\label{fig:task-example}
\end{figure}

Tasks in experiments 1 to 3 are a binary classification of sinusoidal gratings, with each grating defined by its orientation in radians \(\theta\), spatial frequency (s.f.) \(\omega\), and phase \(\phi\).
To set these parameters, $\left[\omega, \theta\right]$ are drawn from the task category distributions $\mathcal{T}$, though $\phi$ is task independent.
Across all curricula, success requires classifying gratings according to their s.f., invariant to orientation and phase.
See \cref{fig:task-example} for an example.

For each task, we first draw $n=1000$ category labels $\mathbf{y} \in \mathbb{B}^{n}$ from a Bernoulli distribution, before drawing parameters for the images themselves from their corresponding category distribution $\mathcal{T}_j$,
\begin{align*}
    y = \mathbf{y}_i &\sim \text{Bern}(0.5) \ , \\
    \begin{bmatrix}\omega_i, \ \theta_i\end{bmatrix}^{\top} &\sim \mathcal{T}_{j,y} \ .
\end{align*}
Equipped with $\omega_i$ and $\theta_i$, the sampled image in terms of its pixel luminosity $l$ (for all pixels $\mathsf{x}$ and $\mathsf{y}$) is, following \citep{Goodfellow2012},
\begin{align*}
    \mathbf{x}_i &= \begin{bmatrix}l(0, 0), \ l(0, 1), \ \hdots, \ l(15, 15) \end{bmatrix}^{\top} \ , \\
    l(\mathsf{x}, \mathsf{y}) &= \sin(\omega_i(\mathsf{x}\cos(\theta_i) + \mathsf{y}\sin(\theta_i) - \phi_i)) \ , 
\end{align*}
where \(\phi_i \sim \mathcal{U}(0, 1)\).
Due to \(\phi\), \(l(\cdot)\) is an embedding from the low-dimensional latent task space to noisy, high-dimensional image space.
Thus, by manipulating the parameters of the tasks' category distributions, we control latent task similarity.

\subsubsection*{\textbf{Experiment 1:} Contrasting curricula}

In our first behavioral experiment, we contrast two different three-task ($N=3$) curricula with fixed parameters.
Both curricula are identical but for the dimension of \(\bm{\mu}\) that varies between tasks.
In the semantic-shift condition \(\mathcal{C}_S\) the s.f.\ dimension of the mean vector varies but the orientation remains fixed. 
Conversely in the perceptual-shift condition \(\mathcal{C}_P\) the s.f.\ dimension has fixed mean but the orientation varies.
See \cref{sec:app-category-parameters} for curricula parameters.

As categories are always \(\omega\)-separable, \(\mathcal{C}_S\) requires learning a new separating plane for each task, without forgetting previous category boundaries.
In contrast, learning \(\mathcal{C}_P\) only requires learning a single category boundary, though invariant to changes in \(\theta\) between tasks.
See \cref{fig:curricula} for example category distributions.

\subsubsection*{\textbf{Experiments 2--3:} Randomized curricula}

Our second and third behavioral experiments introduce the random task paradigm.
Here we generate random curricula, $\mathcal{C}_{Ri}$, comprising pairs of random tasks $\mathcal{T}_{Rj}$,
\begin{align*}
    \mathcal{C}_{Ri} &= \{\: \mathcal{T}_{R1}, \ \mathcal{T}_{R2} \:\} \ , \quad 1 \leq i \leq 30 \ , \\ 
    \mathcal{T}_{Rj} &= \{\: \mathcal{N}(\bm{\mu}_{j}, \bm{\Sigma}) \ , \ \mathcal{N}(\bm{\upsilon}_{j}, \bm{\Sigma})\:\} \ ,
\end{align*}
The mean vectors for the first task $\mathcal{T}_{R1}$ are drawn from a 2D uniform distribution across latent feature space,
\begin{align*}
    \bm{\mu}_{1} &\in \mathcal{U}(0.2, 0.8) \quad \textrm{and} \quad \bm{\upsilon}_{1} = \bm{\mu}_{1} + \mathbf{c} \ ,
\end{align*}
where $\mathbf{c} = \left[0.2, 0\right]^{\top}$ is a fixed vector defining category separation.
The second task $\mathcal{T}_{R2}$ is a modification of the first task by $\mathbf{d}_i$,
\begin{align*}
    \mathbf{d}_i &= \mathcal{U}(-1, 1) \ , \quad \mathbf{\hat{d}}_i = \frac{\mathbf{d}_i}{|\mathbf{d}_i|} \ , \\
    \bm{\mu}_{2} &= \bm{\mu}_{1} + \gamma \mathbf{\hat{d}}_i \quad \textrm{and} \quad \bm{\upsilon}_{2} = \bm{\upsilon}_{1} + \gamma \mathbf{\hat{d}}_i \ ,
\end{align*}
where $\gamma = 0.25$ is a scaling constant that fixes the distance between tasks (see \cref{sec:app-effect-gamma} for different choices of $\gamma$).
According to $\mathbf{c}$ and $\gamma$, both the inter-task and inter-category distances are held constant.
Unlike experiment 1, task pairs can differ in both the semantic and perceptual dimensions, depending on the direction of $\mathbf{d}_i$.
In experiment 2, the first task is held constant and the second task is randomized.
In experiment 3, both tasks are drawn at random.

\subsection{Naturalistic experiments}

\subsubsection*{\textbf{Experiment 4:} Randomized MNIST}

Our fourth experiment tests the validity of our findings beyond the synthetic gratings task.
We turn to a randomized variant of split MNIST \citep{Zenke2017},
\begin{align*}
    \mathcal{C}_{Mi} &= \{\: \mathcal{T}_{M1}, \ \mathcal{T}_{M2} \:\} \ , \quad 1 \leq i \leq 30 \ , \\ 
    \mathcal{T}_{Mj} &= \{\: (X_{j0}, \ y_{j0}) \ , \ (X_{j1}, \ y_{j1}) \:\} \ ,
\end{align*}
where $y_{j0}$, $y_{j1}$ are sampled without replacement from the class labels $\{0, \ldots, 9\}$, and $X_{j0}$, $X_{j1}$ are the corresponding inputs.
For consistency with experiments 1--3, input images are resized to $16\times16$ pixels using bilinear interpolation.
The first task in every curriculum is a binary classification of a random pair of digits, and the second task is another random pair.

\subsubsection*{\textbf{Experiment 5:} ResNet with randomized CIFAR}

Our final experiment tests the whether our claims generalize to more a more complex model.
We train ResNet-18 \citep{He2016} on a randomized variant of split CIFAR-10 \citep{Krizhevsky2009}.
As with experiment 4, we sample 30 binary classification tasks, though here using the original $32\times32$ color images.
See \cref{sec:app-category-parameters} for full task parameters for all five experiments.

\subsection{Model architectures}\label{sec:model-arch}

Experiments 1 to 4 are performed on two-layer ReLU networks.
Let $\mathbf{x} \in \mathbb{R}^{d}$ be an input vector representing an image, and let $W_1 \in \mathbb{R}^{d \times h}$, $W_2 \in \mathbb{R}^{h \times 1}$ and $\mathbf{b}_1 \in \mathbb{R}^{h}$, $\mathbf{b}_2 \in \mathbb{R}^{1}$ be the weights and biases for the first and second layers.
Then, the network's decision $\hat{y} \in \mathbb{R}$ is
\begin{align*}
    \hat{y} &= s(W_2^{\top} \mathbf{a} + \mathbf{b}_2) \ , \quad \mathbf{a} = r(W_1^{\top} \mathbf{x} + \mathbf{b}_1) \ ,
\end{align*}
where
\begin{align*}
    r(x) &= \max(0,x) \ , \quad s(x) = \frac{1}{1 + e^{-x}} \ ,
\end{align*}
i.e.\ $r(\cdot)$ and $s(\cdot)$ are the element-wise ReLU and sigmoid functions respectively.
For our specific networks, $d = h = 256$, and batch normalization is used to center and scale the inputs to have zero mean and unit variance.
Model parameters are randomly initialized from $\mathcal{U}(-\sqrt{d}, \sqrt{d})$.

In experiment 5, we use the Torchvision \citep{Paszke2017} implementation of ResNet-18, modified to have a scalar output.
Model parameters are initialized according to \citet{He2016}.

We note that continual learning researchers often turn to ``multi-headed'' models, where a new output layer is added for each task.
As this requires extra parameters for every task, alongside explicit direction to use a specific output, it presents a rather limited scenario given boundaries between tasks are not always well-defined \citep{Farquhar2018}.
In contrast, we opt for the easier-to-understand yet arguably more challenging ``single-headed'' setting, using a single output layer and without information about task changeover.
We suggest that solid empirical foundations in this simpler setting are a necessary prerequisite for robust future research in continual learning.

\subsection{Model training}\label{sec:model-training}

Model training is the same across all curricula in experiments 1 to 5 unless otherwise noted.
Parameters are randomly initialized at the start of each curriculum, and for subsequent tasks model parameters for task $j$ are initialized to $\bm{\theta}_{j-1}$.
The network is then trained sequentially on each task, using batches of 128 images, for 1,000 epochs in experiments 1, 2, 3 and 5, and 5,000 epochs in experiment 4.
Networks are trained to minimize the cross-entropy loss using stochastic gradient descent with learning rate 0.001, with neither learning rate decay nor explicit regularization penalties.
10 models with different random seeds are trained for each condition in experiments 1 to 4, with 5 models trained for experiment 5 due to compute constraints.
Number of epochs and learning rate were manually set according to pilot experiments.
No other model or optimization hyperparameters were tuned, though see \cref{sec:app-effect-lr,sec:app-effect-l2} for a post-hoc analysis of the effect of learning rate and weight decay.

\subsection{Loss surface analysis}\label{sec:methods-grads}

Using the population of models trained in our behavioral experiments, we now investigate the loss function gradients with the goal of identifying how they relate to CF\@.
\citet{Mirzadeh2020} argues that the curvature of $L_{j}(\bm{\theta}_{j})$ determines forgetting, loosely bounding the change in first task loss between $\bm{\theta}_{j}$ and $\bm{\theta}_{j+1}$ by the spectral radius of the Hessian of the first task loss.
In \cref{sec:results-task-1-fixed} we present behavioral evidence that this only partially explains CF, showing that from a constant first task (i.e.\ assuming a constant minima) CF varies significantly depending on the choice of \emph{second} task.
Thus, we suggest that the first gradient update of the \emph{new} task can also explain CF.

Like \citeauthor{Mirzadeh2020}, we approximate the loss surface in with a second-order Taylor expansion, though we focus on $L_{j+1}$ rather than $L_{j}$:
\begin{align}\label{eq:taylor}
    L_{j+1}(\bm{\theta}_{j} + \epsilon \mathbf{g}) = L_{j+1}(\bm{\theta}_{j}) + \epsilon \mathbf{g}^{\top}\mathbf{g} + \frac{1}{2} \epsilon^2 \mathbf{g}^{\top} H \mathbf{g},
\end{align}
where $\mathbf{g}=\nabla_{\theta_{j}}L_{j+1}(\bm{\theta}_{j})$ is the gradient and $H=\nabla^{2}_{\theta_{j}}L_{j+1}(\bm{\theta}_{j})$ is the Hessian matrix around $\bm{\theta}_{j}$, the collection of $(W_1, W_2, \mathbf{b}_1, \mathbf{b}_2)$ after training on task $j$.

We note that most prior work makes use of the difference between the post-training minima of the two tasks, predicating analysis on completing second task training before determining whether CF will occur.
In contrast, using the fully randomized curricula of experiments 3--5, we show that a single gradient update---calculated using a single backward pass---is sufficient.
From \cref{eq:taylor}, the change in loss is determined by the norm of the gradient ($\mathbf{g}^{\top}\mathbf{g}$), corrected for curvature ($\mathbf{g}^{\top}H\mathbf{g}$).
As computing $H$ is impractical for a large number of parameters, we crudely approximate $H$ using the diagonal of the empirical Fisher information matrix \citep{Ritter2018, Immer2021, Kirkpatrick2017},
\begin{align*}
    H \approx F &= \sum_i \nabla_{\theta_{j}} (\log P(y_i|\mathbf{x}_i;\bm{\theta}_{j}))(\log P(y_i|\mathbf{x}_i;\bm{\theta}_{j})^{\top}) \\ 
                &= \mathbf{g}\mathbf{g}^{\top} \\
    			&\approx \text{Diag}(\text{diag}(\mathbf{g}\mathbf{g}^{\top})) \\
                &= \text{Diag}(\mathbf{g}^2) \ ,
\end{align*}
where $\text{Diag}(\mathbf{v})$ returns a diagonal matrix, and $\text{diag}(M)$ returns its diagonal elements.
As such we roughly approximate the dominant eigenvalue of $H$ with $\|\mathbf{g}^{2}\|_{\infty}$, providing an alternative to the $\lambda_{max}$ bound of \citeauthor{Mirzadeh2020}.
In \cref{sec:grad-predict}, we show that even the first step away from $\bm{\theta}_j$ determines CF, and as such $\|\mathbf{g}\|_{\infty}$ is strongly correlated with future forgetting. 


\section{Results}

\begin{figure}[h]
    \centering
    \includegraphics[width=0.38\textwidth]{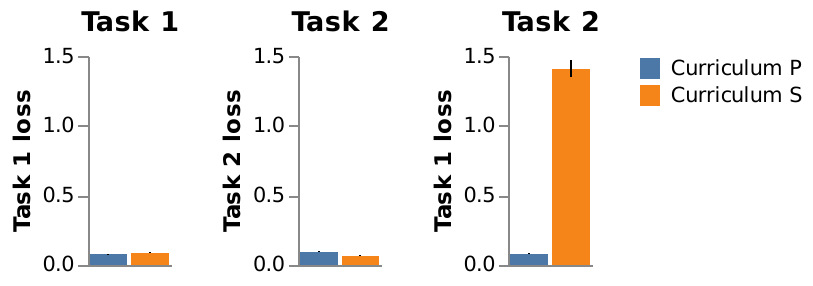}
    \caption{\textbf{Left:} Mean task 1 validation loss after training on task 1 in experiment 1. \textbf{Middle:} Task 2 loss after tasks 1 and 2. \textbf{Right:} Task 1 loss after tasks 1 and 2. Only $\mathcal{C}_S$  exhibits CF. Bars are STD.}\label{fig:loss-end}
\end{figure}

\begin{figure*}
     \centering
     \includegraphics[width=0.75\textwidth]{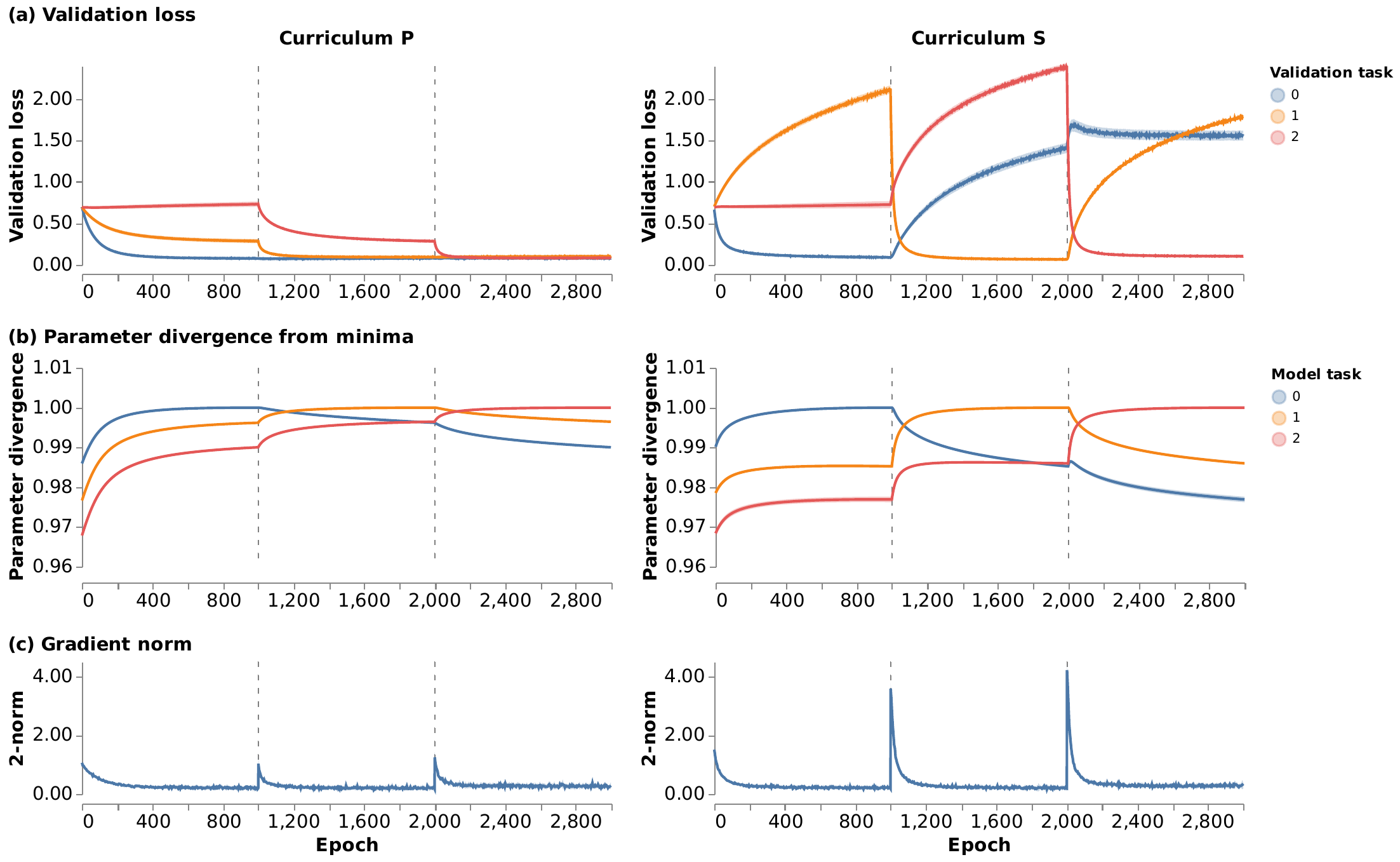}
     \caption{\textbf{Left:} $\mathcal{C}_P$. \textbf{Right:} $\mathcal{C}_S$. \textbf{(a)} Mean validation loss for all three tasks with sequential training from task to task. Grey dashed lines indicate task change. Only $\mathcal{C}_S$ exhibits CF. \textbf{(b)} Dot product of model parameters at each epoch and at minima for each task. Parameters sharply diverge from minima in $\mathcal{C}_S$. \textbf{(c)} 2-norm of gradient of training loss w.r.t.\ all parameters. Larger gradient updates take place in $\mathcal{C}_S$. Shaded region (barely visible) is STD.}\label{fig:loss-and-grads-continuous}

    \centering
    \subfloat[Gratings (first task fixed)]{
        \includegraphics[width=0.38\textwidth]{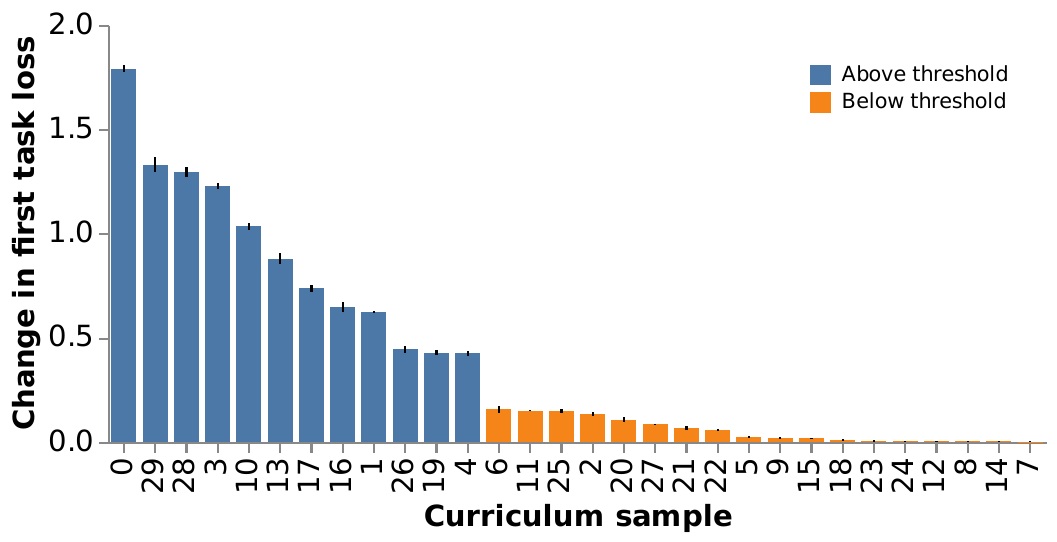}
    }
    \subfloat[Gratings (both tasks random)]{
        \includegraphics[width=0.38\textwidth]{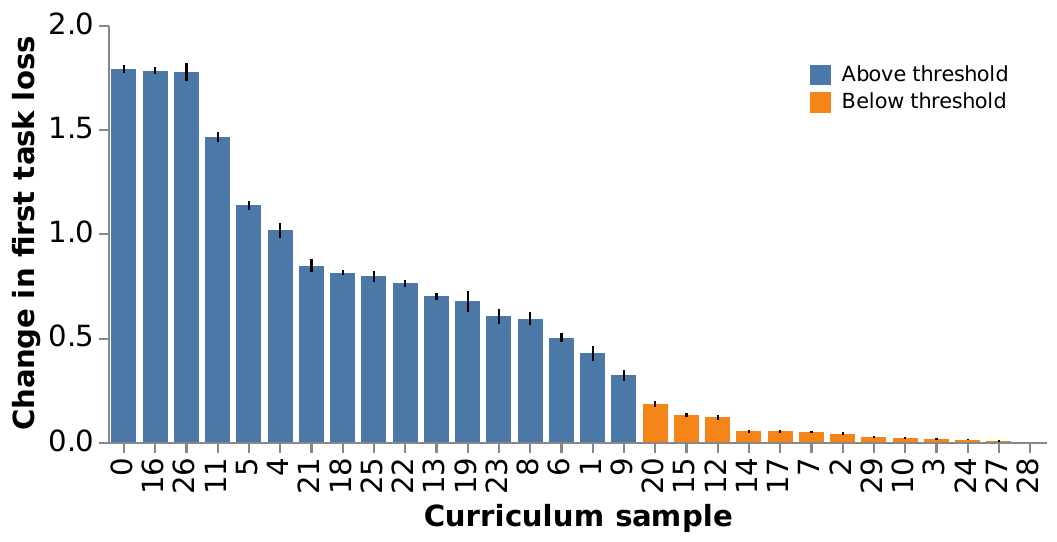}
    }

    \subfloat[ReLU network on MNIST]{
        \includegraphics[width=0.38\textwidth]{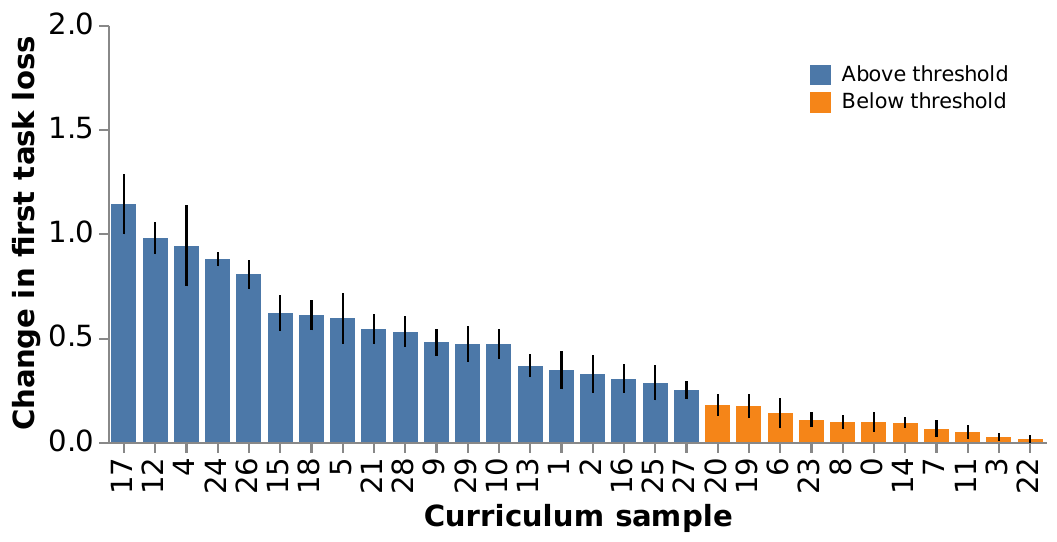}
    }
    \subfloat[ResNet-18 on CIFAR-10]{
        \includegraphics[width=0.38\textwidth]{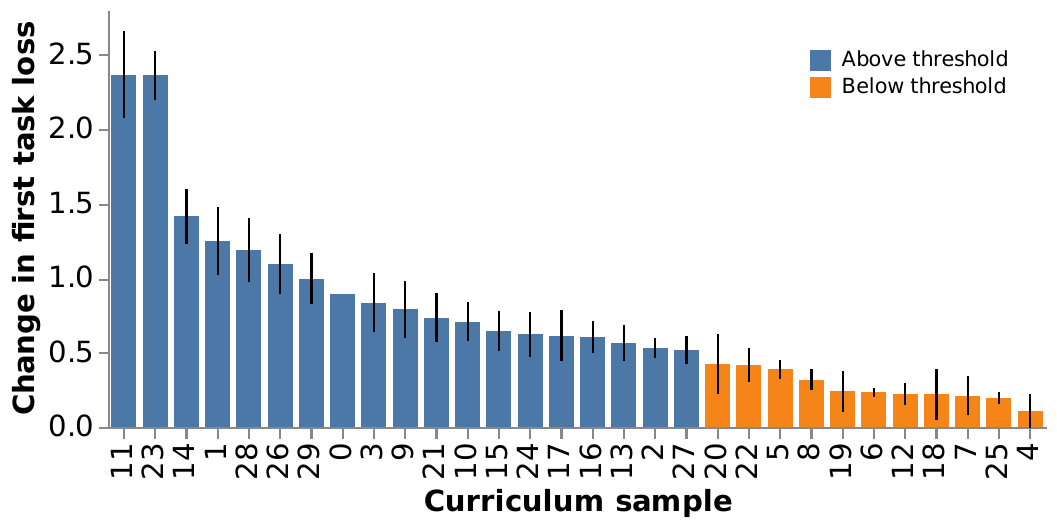}
    }
    \caption{Change in mean first task validation loss after training on second task over curricula of \textbf{(a)} gratings, first task fixed and second randomized; \textbf{(b)} gratings, both tasks randomized; \textbf{(c)} split MNIST, both tasks randomized. \textbf{(d)} is ResNet-18 trained on split CIFAR-10, note different scale and higher threshold \(\epsilon=0.5\). Only certain curricula exhibit forgetting in all experiments. Bars are STD.}\label{fig:randomised-loss}
\end{figure*}



\begin{figure*}[h]
    \centering
    \subfloat[Gratings]{
        \includegraphics[width=0.29\textwidth]{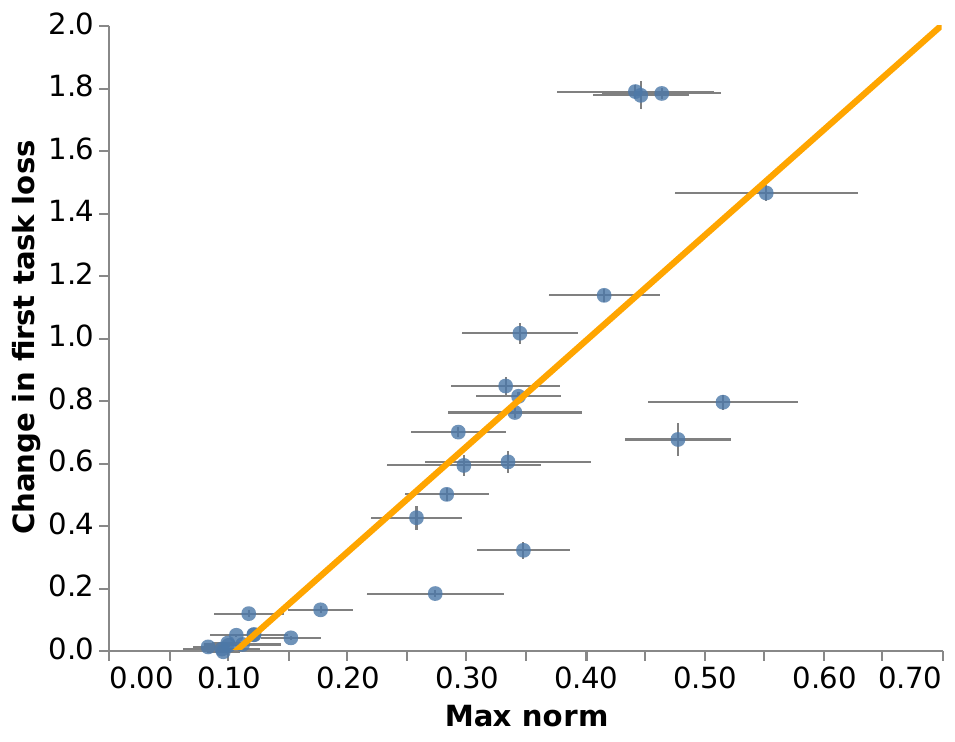}
    }
    \subfloat[ReLU network on MNIST]{
        \includegraphics[width=0.29\textwidth]{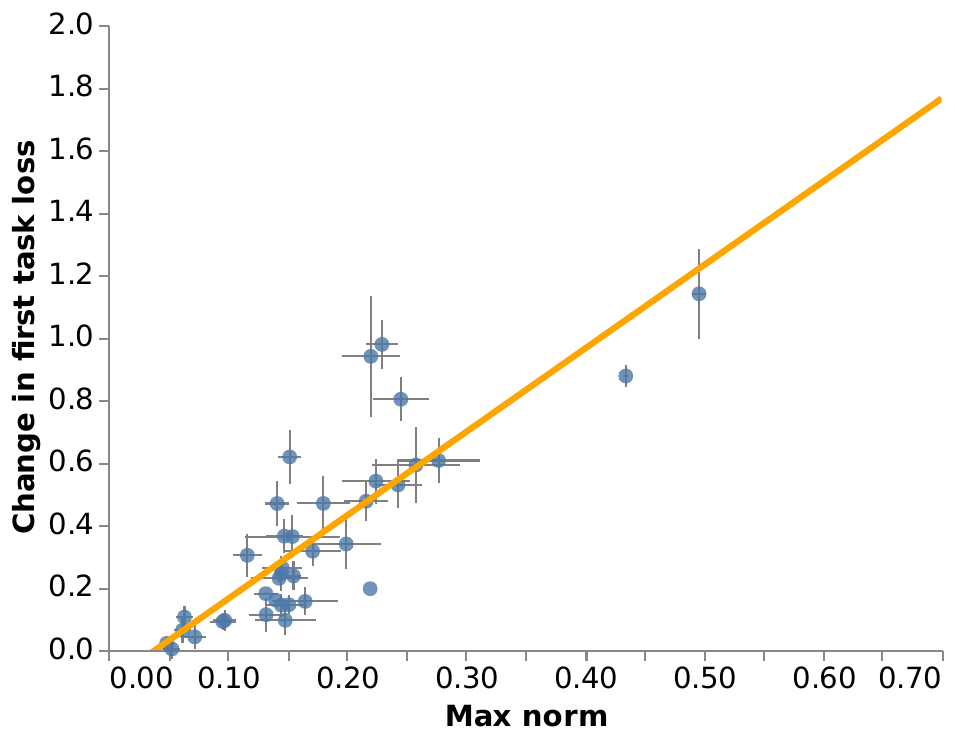}
    }
    \subfloat[ResNet-18 on CIFAR-10]{
        \includegraphics[width=0.29\textwidth]{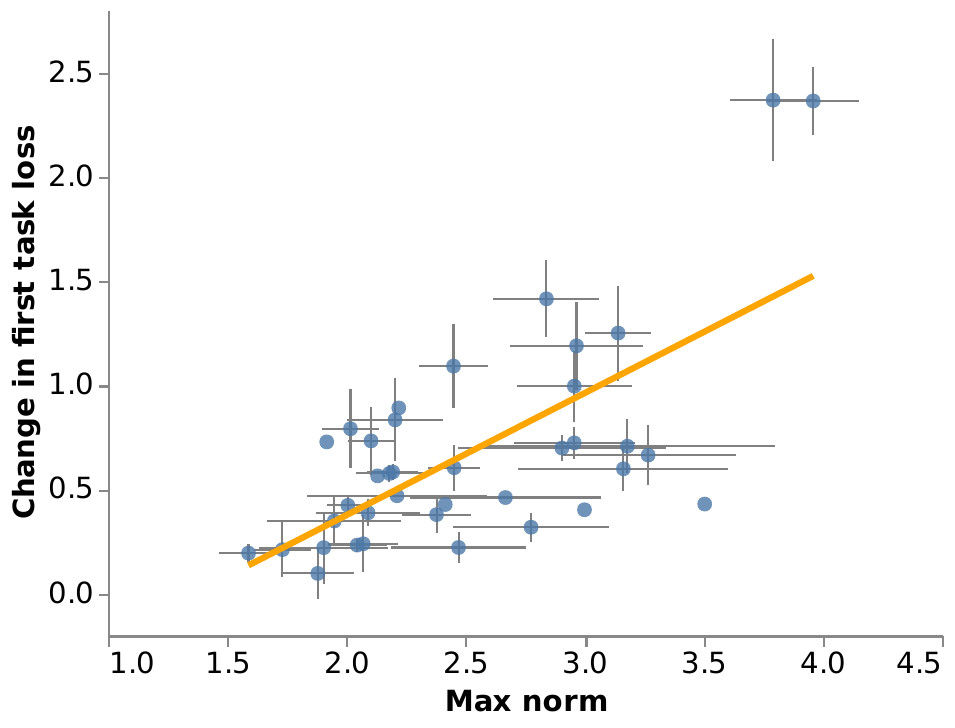}
    }
    \caption{Change in first task loss and $\infty$-norm of gradient of second task training loss after single pass on second task. In both \textbf{(a)} randomized curricula and \textbf{(b)} split MNIST the $\infty$-norm is strongly positively and highly significantly correlated with CF\@. \textbf{(c)} ResNet-18 on CIFAR-10 exhibits the same correlation; note different scale. Orange regression line is for illustration only. Bars are STD.}\label{fig:loss-and-grad-norm}
\end{figure*}

\subsection{Semantic task difference drives CF}

Experiment 1 demonstrates that CF only occurs in certain situations (\cref{fig:loss-end}).
In both conditions models learn to perform both tasks, achieving validation loss $L_{1}(\bm{\theta}_1)$ of $0.087\pm0.002$ (mean $\pm$ SD) and $0.076\pm0.002$ for $\mathcal{C}_S$ and $\mathcal{C}_P$ respectively, and $L_{2}(\bm{\theta}_{2})$ of $0.064\pm0.004$ and $0.094\pm0.003$.
CF is selectively apparent, with $L_{1}(\bm{\theta}_2)$ rising significantly to $1.412\pm0.060$ for $\mathcal{C}_S$ and remaining at $0.079\pm0.005$ for $\mathcal{C}_P$ ($t(18) = 70.18$, $p < .0001$, Welch's $t$-test for unequal variances).
This suggests CF is most significant when tasks are perceptually similar but vary in their semantic mapping aligning with results presented by \citet{Ramasesh2020} and \citet{Lee2021}.

\subsection{Parameter dynamics mirror CF}

\Cref{fig:loss-and-grads-continuous}{a} shows the validation losses for each task in both curricula of experiment 1 over time.
In both curricula, as training on the $j$th task commences, $L_{j}(\cdot)$ decreases until converging to its minimum.
In curriculum $\mathcal{C}_P$, once a model has reached its minimum loss for task $j$, it is maintained for all tasks $k > j$. 
In curriculum $\mathcal{C}_S$, however, loss $L_{j}(\cdot)$ sharply increases as soon as training progresses onto subsequent tasks $k > j$.
These dynamics are broadly mirrored by network parameter change as training progresses across tasks.
\Cref{fig:loss-and-grads-continuous}{b} shows the deviation of the parameters at time $t$ from the minima reached at the end of training on task $j$, i.e.\ $\bm{\theta}_{j}^{\top}\bm{\theta}^{(t)}$.
In $\mathcal{C}_S$, parameters move rapidly away from their minima as training crosses the task boundary.
\Cref{fig:loss-and-grads-continuous}{c} shows the 2-norm of the gradient vector through training across tasks, demonstrating much sharper peaks at the task boundaries in $\mathcal{C}_S$ than $\mathcal{C}_P$.
In \cref{sec:grad-predict}, we show that the size of this first update after the boundary is predictive of CF\@.

\subsection{First task curvature}\label{sec:results-task-1-fixed}

\citet{Mirzadeh2020} argue that the curvature of the first task loss surface around the achieved minima determines forgetting, suggesting that a shallower curvature leads to less forgetting. 
In experiment 2, we hold the first task constant and randomly vary the second task, yielding 30 different curricula.
\Cref{fig:randomised-loss}{a} shows that from this constant first task, using a threshold of $\epsilon = 0.2$, only 12 out of 30 curricula result in CF, with first task loss $L_{1}(\bm{\theta}_{2})$ ranging from $0.063$ to $1.891$.
We assume that as we hold the first task, and hence its loss function, fixed, that the curvature of the obtained minima is also constant.
Thus, from a simple behavioral experiment, we suggest that first task curvature can only partially explain forgetting.

\subsection{Random curricula}

\begin{figure}[h]
    \centering
    \subfloat[Effect of semantic similarity]{
        \includegraphics[trim={0 0 0 0.24cm},clip,width=0.29\textwidth]{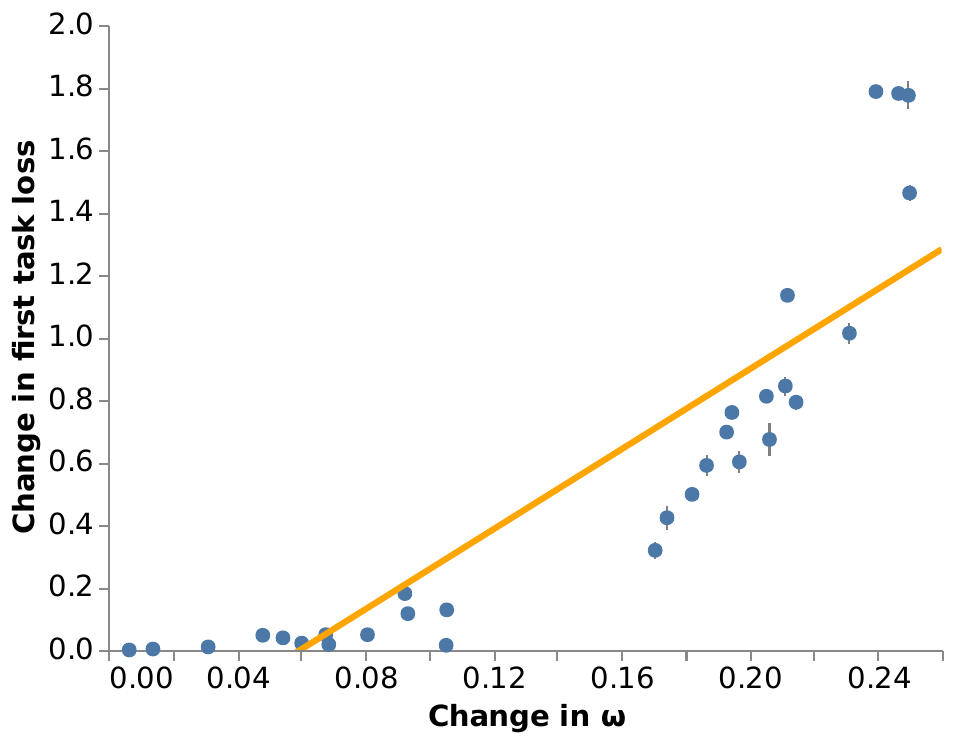}
    }
    
    \subfloat[Effect of perceptual similarity]{
        \includegraphics[trim={0 0 0 0.24cm},clip,width=0.29\textwidth]{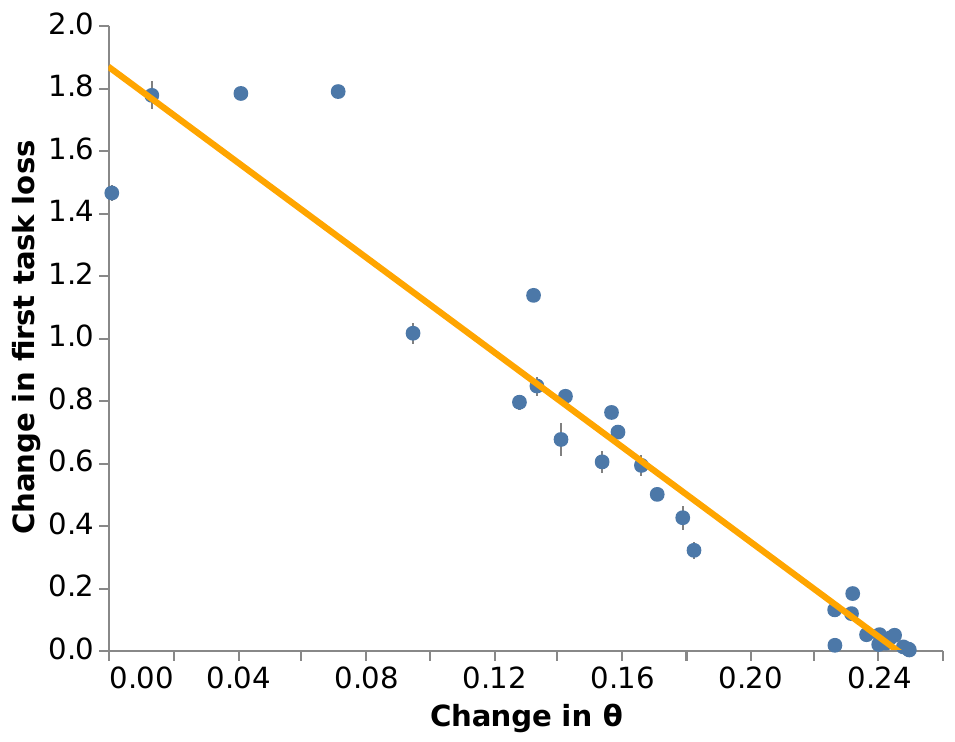}
    }
    \caption{\textbf{(a)} Semantic distance leads to greater CF\@. \textbf{(b)} Perceptual distance leads to reduced CF\@. Orange regression line is for illustration only. Bars are STD.}\label{fig:loss-perceptual-semantic}
\end{figure}

In experiment 3, all models successfully learn to perform both tasks in order, with $L_{1}(\bm{\theta}_{1}) = 0.075\pm0.011$ and $L_{2}(\bm{\theta}_{2}) = 0.069\pm0.015$.
After training on the second task, first task loss $L_{1}(\bm{\theta}_{2})$ rises to $0.631\pm0.567$, with one model reaching a minimum loss of $0.052$ and another reaching a maximum of $1.921$.
Only certain task pairs (17/30) induce CF (\cref{fig:randomised-loss}{b}).

\Cref{fig:loss-perceptual-semantic} shows the effect of task distance along either the semantic or perceptual dimension.
A semantic shift between tasks is associated with increased CF (Pearson's $r(28) = -0.97$, $p < .0001$), whereas a perceptual shift is associated with a decrease ($r(28) = 0.88$, $p < .0001$).

We see similar results in our naturalistic experiments.
In experiment 4, all ReLU network models successfully learn both tasks, reaching $L_{1}(\bm{\theta}_{1}) = 0.062\pm0.032$ and $L_{2}(\bm{\theta}_{2}) = 0.067\pm0.034$.
First task loss after training on task 2 is $L_{1}(\bm{\theta}_{2}) = 0.565\pm0.758$, with a minimum of $0.006$ and a maximum of $3.905$, and 18/30 curricula exhibiting forgetting (\cref{fig:randomised-loss}{c}). 
In experiment 5, the ResNet models also learn both tasks though with slightly higher final validation loss ($L_{1}(\bm{\theta}_{1}) = 0.311\pm0.165$; $L_{2}(\bm{\theta}_{2}) = 0.367\pm0.191$).
First task loss after training on task 2 is $L_{1}(\bm{\theta}_{2}) = 1.030\pm0.588$ (min $0.272$; max $2.895$), with 19/30 curricula exhibiting forgetting (\cref{fig:randomised-loss}{d}).
Note we use $\epsilon=0.5$ for the ResNet experiments in response to higher loss at the end of training.

\subsection{First update gradient norm predicts CF}\label{sec:grad-predict}

Following \cref{sec:methods-grads}, we present evidence that the max-norm of the first gradient step of the \emph{new} task, $\|\nabla_{\theta_{j}}L_{j+1}(\bm{\theta}_{j})\|_{\infty}$, is predictive of forgetting.
For both experiments 3 and 4, over gratings and split MNIST respectively  (see \cref{fig:loss-and-grad-norm}{a,b}), the change in first task loss is strongly positively and highly significantly correlated with the max-norm of the new task first gradient step (exp.\ 3: Pearson's $r(28) = 0.87$, $p < .0001$; exp.\ 4: $r(28) = 0.83$, $p < .0001$).
In experiment 5 with ResNet-18 on CIFAR-10 (\cref{fig:loss-and-grad-norm}{c}), $\|\nabla_{\theta_{j}}L_{j+1}(\bm{\theta}_{j})\|_{\infty}$ remains positively and highly-significantly correlated with forgetting (exp.\ 5: $r(28) = 0.67$, $p < .0001$).
Supporting our focus on the new task, the max norm of an additional gradient step minimizing the original loss, $\|\nabla_{\theta_{j}}L_{j}(\bm{\theta}_{j})\|_{\infty}$, is uncorrelated with the change in loss (exp.\ 3: $r(28) = -0.01$, $p = .96$; exp.\ 4: $r(28) = 0.04$, $p = .82$; exp.\ 5: $r(28) = -0.11$, $p = .49$).


\section{Discussion}\label{sec:discussion}

This study demonstrates how carefully-designed experiments can elucidate neural network behavior.
Here, our  case study phenomenon is catastrophic forgetting, and through our experiments we have presented a number of behavioral findings.
First, we have shown that CF is associated more with semantic task differences than perceptual, aligning with previous work \citep{Ramasesh2020, Lee2021}.
Second, we have demonstrated that the previous task loss surface only partially accounts for CF, and that the new task surface is also an important factor.
Taken together, these results paint a fine-grained picture of CF, and suggest that closer investigation is required on the exact nature of the inter-task relationship.
Beyond behavior, our results have led us to a simple and easy-to-compute heuristic, the max-norm of the first gradient step, that can immediately predict the occurrence of CF.

Fields such as statistical mechanics provide a useful bottom-up framework for reasoning about neural networks as complex systems in terms of their parts.
However, as increasingly complex models give rise to emergent and unpredictable phenomena, they may call for new approaches without the associated reductionism.
Even equipped with full knowledge of a model and its parameters, this is unlikely sufficient to predict whether and how a model will learn, generalize, transfer, forget or any number of other complex behaviors.
In this work, we have argued that for this level of understanding, we need the fundamental tool of experimental psychology: the behavioral experiment.

We foresee opportunities for targeted behavioral experiments across our discipline, but particularly in two scenarios. 
First, when investigating other surprising phenomena in large neural networks, for example generalization in the overparameterized regime \citep{Belkin2019, Nakkiran2019}; scaling laws \citep{Tan2019, Kaplan2020, Hernandez2021}; and learning dynamics \citep{Power2022}.
Second, when evaluating bias in applied machine learning systems \citep{Dixon2018, Buolamwini2018}, which when unchecked can lead to significant and damaging personal consequences.
Further behavioral experiments can both aid our general understanding and help to guard our systems against unintended function.

\section*{Acknowledgments}
We are grateful to Onno Kampman and the anonymous reviewers for their helpful feedback.
SB is supported by the Biotechnology and Biological Sciences Research Council [grant number BBSRC BB/M011194/1].
NL is supported by a Senior Turing AI Fellowship funded by the UK government’s Office for AI, through UK Research and Innovation (grant reference EP/V030302/1), and delivered by the Alan Turing Institute.
NL's chair is endowed by DeepMind.



\appendix

\section*{Appendix}

\begin{figure*} 
    \centering
    \subfloat{
        \includegraphics[height=5cm]{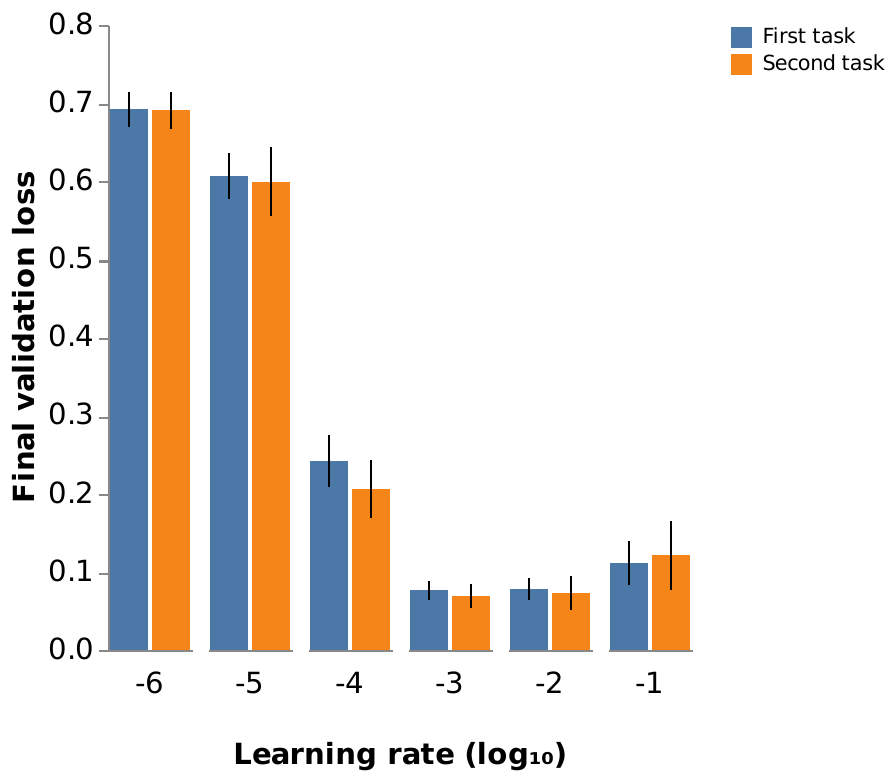}
        \label{fig:final-loss-by-lr}
    }
    \subfloat{
        \includegraphics[height=5cm]{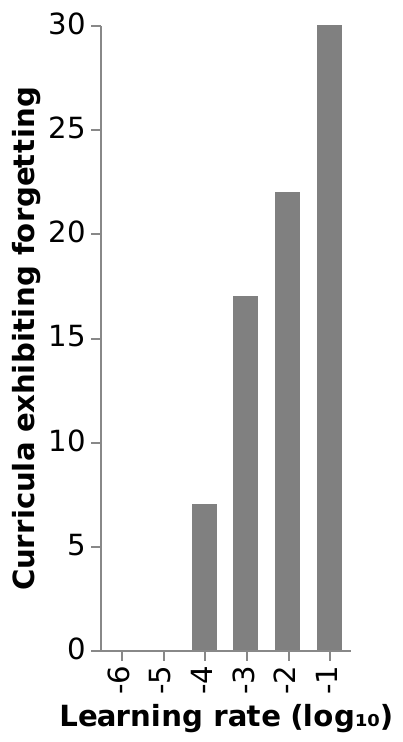}
        \label{fig:num-forgetting-by-lr}
    }
    \subfloat{
        \includegraphics[height=5cm]{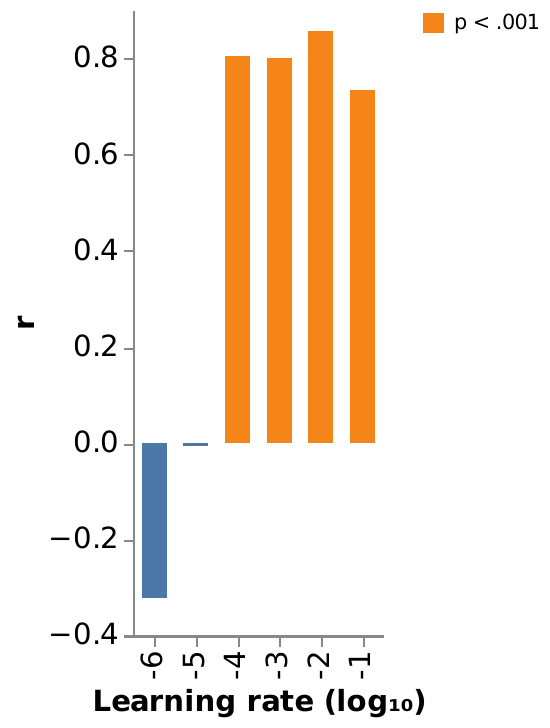}
        \label{fig:pearson-by-lr}
    }
    \caption{\textbf{Experiment 3:} \textbf{(a)} Final validation loss on each task by learning rate. \textbf{(b)} Number of curricula exhibiting forgetting by learning rate. \textbf{(c)} Pearson correlation of change in first task loss and max-norm of first gradient update, by learning rate.}\label{fig:lr-breakdown}

\end{figure*}

\section{Curricula category parameters}\label{sec:app-category-parameters}

Task parameters for both curricula in experiment 1 can be found in \cref{table:manual}.
Each task is a binary classification of two 2D Gaussian categories, with $\mu$ and $\upsilon$ as mean vectors for each category distribution.
All categories have covariance $\sigma^{2}I$ with $\sigma^2 = 0.05$.

\Cref{table:randomised} details task parameters for each random curriculum in experiments 2 and 3. 
In experiment 2, the first task is always equal to the first task of curriculum 0.
In experiment 3, both first and second task parameters are as described in the table.

\Cref{table:mnist} details digit pairings for each random split MNIST curriculum in experiment 4, and \Cref{table:cifar} details class pairings for split CIFAR10 in experiment 5.

\section{Effect of learning rate}\label{sec:app-effect-lr}

In the main text learning rate is fixed at 0.001.
Here, we present a supplementary analysis of different choices of learning rate on our principal results from experiment 3.
\Cref{fig:lr-breakdown}{a} shows the first and second task validation loss after training on each task, indicating that the task is not successfully learned for learning rates $\leq$\SI{1e-5}.
For all other learning rates $>$\SI{1e-5}, \cref{fig:lr-breakdown}{b} outlines a clear positive trend: a larger learning rate leads to a larger number of curricula exhibiting forgetting.
\Cref{fig:lr-breakdown}{c} confirms that our max-norm heuristic remains highly-significantly and strongly positively-correlated ($p < .001$) with future catastrophic forgetting in all cases where the tasks have successfully been learned, i.e. LR $>$\SI{1e-5}.

\section{Effect of weight decay}\label{sec:app-effect-l2}

\begin{figure*}
    \centering
    \subfloat{
        \includegraphics[height=5cm]{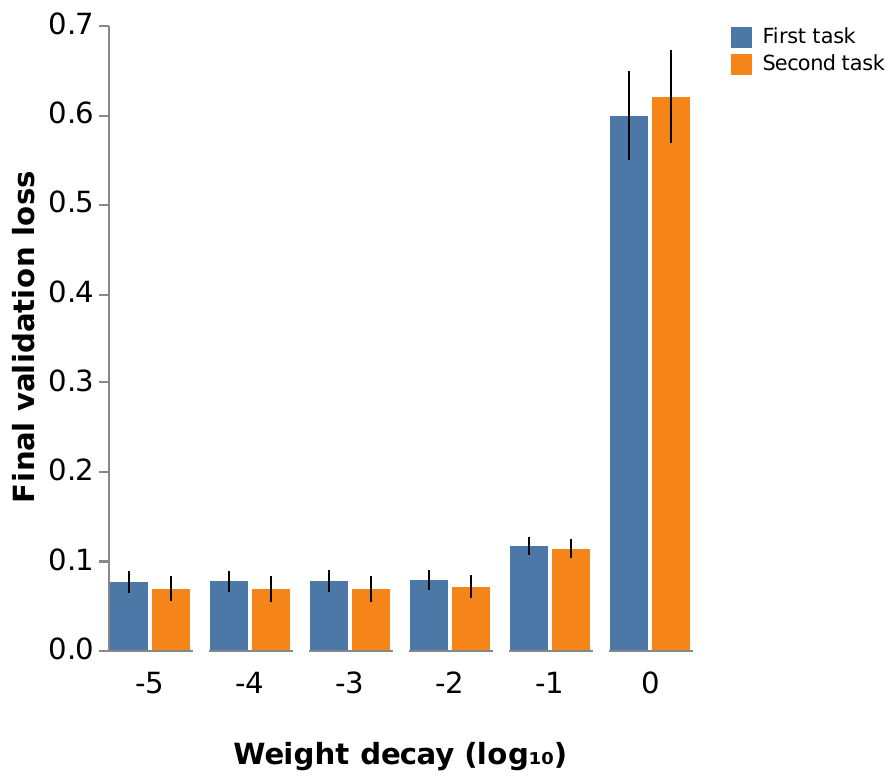}
        \label{fig:final-loss-by-l2}
    }
    \subfloat{
        \includegraphics[height=5cm]{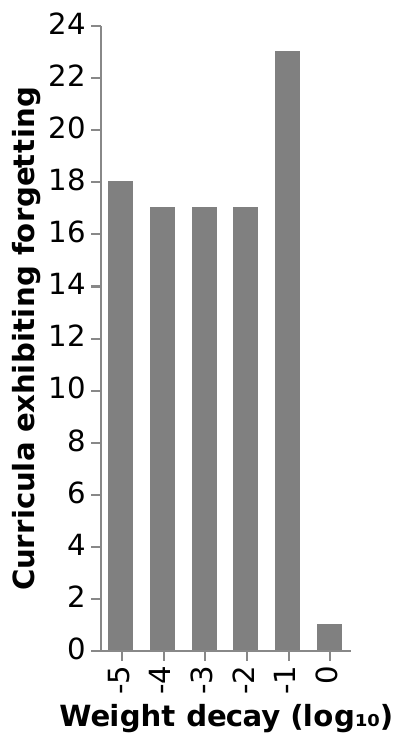}
        \label{fig:num-forgetting-by-l2}
    }
    \subfloat{
        \includegraphics[height=5cm]{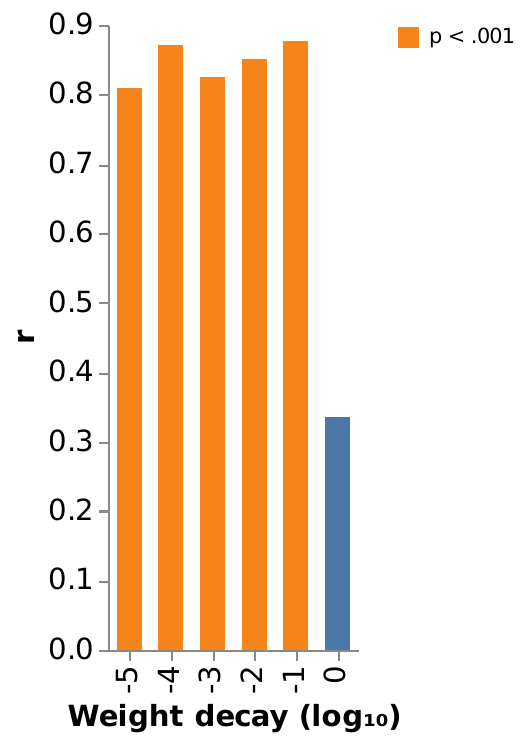}
        \label{fig:pearson-by-l2}
    }
    \caption{\textbf{Experiment 3:} \textbf{(a)} Final validation loss on each task by weight decay. \textbf{(b)} Number of curricula exhibiting forgetting by weight decay. \textbf{(c)} Pearson correlation of change in first task loss and max-norm of first gradient update, by weight decay.}\label{fig:l2-breakdown}

\end{figure*}

Similarly, while our primary experiments do not use weight decay, we here present a supplementary analysis of the effect of weight decay on experiment 3.
From \cref{fig:l2-breakdown}{a} we see that all models trained with all weight decay settings $< 1.0$ successfully learn the task.
In \cref{fig:l2-breakdown}{b} we observe broadly similar numbers of curricula that induce forgetting across models trained on all weight decay values, with an exception for a weight decay of \num{1.0}, where low incidence of CF is best explained by poor performance on the first task even before moving to the second.
We also confirm, in \cref{fig:l2-breakdown}{c}, that the max-norm heuristic remains significantly correlated ($p < .001$) with CF in all scenarios where the first task has been successfully learned.

\section{Effect of task distance}\label{sec:app-effect-gamma}

\begin{figure*}
    \centering
    \subfloat{
        \includegraphics[height=5cm]{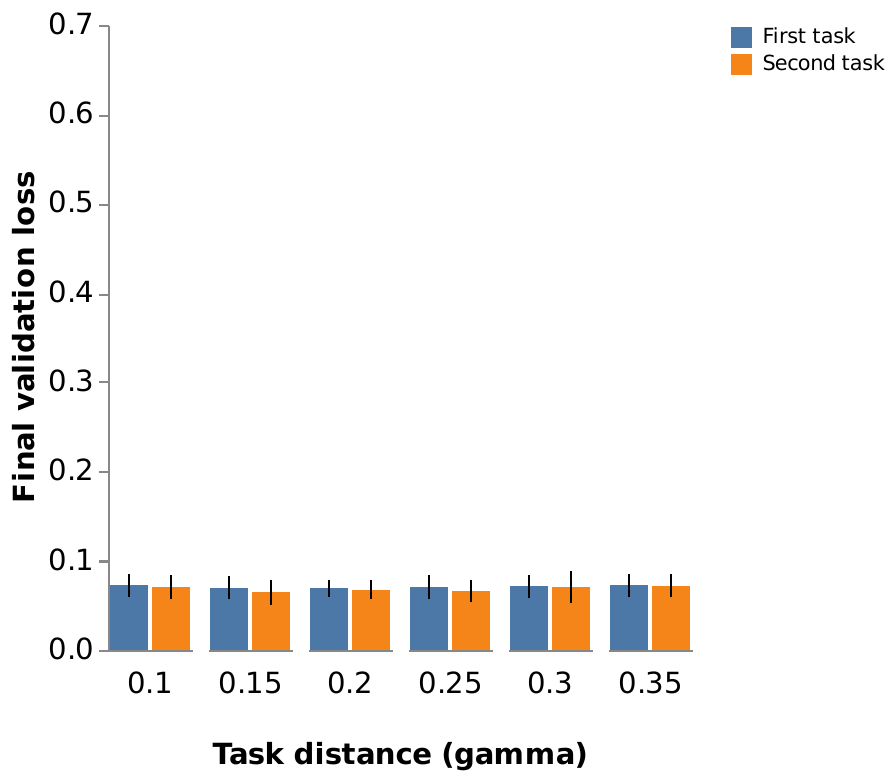}
        \label{fig:final-loss-by-gamma}
    }
    \subfloat{
        \includegraphics[height=5cm]{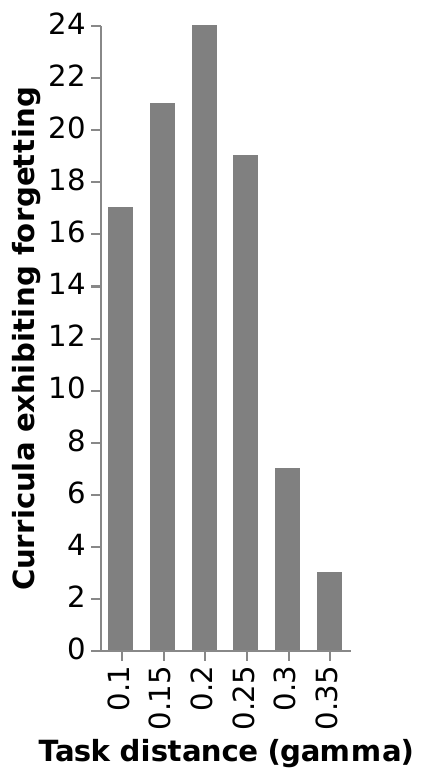}
        \label{fig:num-forgetting-by-gamma}
    }
    \subfloat{
        \includegraphics[height=5cm]{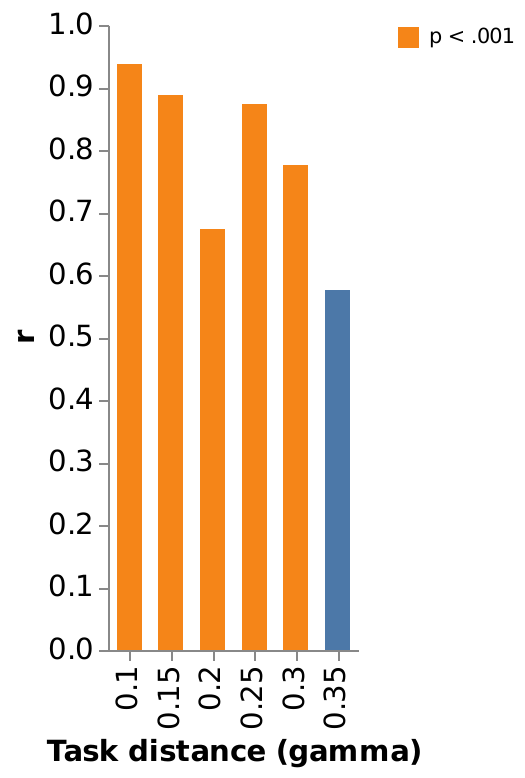}
        \label{fig:pearson-by-gamma}
    }
    \caption{\textbf{Experiment 3:} \textbf{(a)} Final validation loss on each task by task distance, \(\gamma\). \textbf{(b)} Number of curricula exhibiting forgetting by task distance. \textbf{(c)} Pearson correlation of change in first task loss and max-norm of first gradient update, by task distance.}\label{fig:gamma-breakdown}

\end{figure*}

In our final supplementary analysis, we evaluate the effect of task distance, $\gamma$, i.e.\ a scalar controlling the similarity of two tasks within a curriculum.
\Cref{fig:gamma-breakdown}{a} shows that task distance has no effect on the ability of the models to learn both first and second task in experiment 3.
Interestingly, \cref{fig:gamma-breakdown}{b} suggests that both decreasing and increasing task distance may decrease incidence of CF, though an ``in-between'' task distance (around $\gamma = 0.25$ as used in the main text) is associated with the most CF.
Finally, we note in \cref{fig:gamma-breakdown}{c} that the max-norm heuristic is also strongly and significantly correlated ($p < .001$) in all cases except for largest task distance evaluated here, $\gamma = 0.35$, which remains positively correlated and significant, though at a lower threshold after correction for multiple comparisons ($r = 0.576$, $p = .005$).

\vfill

\begin{table*}
    \centering
    \begin{tabular}{lrrrrrrrrrrrr}
    \toprule
    {} & \multicolumn{4}{l}{Task 1} & \multicolumn{4}{l}{Task 2} & \multicolumn{4}{l}{Task 3} \\
    {} &   $\mu_0$ &  $\mu_1$ & $\upsilon_0$ & $\upsilon_1$ &   $\mu_0$ &  $\mu_1$ & $\upsilon_0$ & $\upsilon_1$ &   $\mu_0$ &  $\mu_1$ & $\upsilon_0$ & $\upsilon_1$ \\
    Curriculum &        &       &           &           &        &       &           &           &        &       &           &           \\
    \midrule
S          &  0.1 & 0.5 &     0.3 &     0.5 &  0.4 & 0.5 &     0.6 &     0.5 &  0.7 & 0.5 &     0.9 &     0.5 \\
P          &  0.4 & 0.2 &     0.6 &     0.2 &  0.4 & 0.4 &     0.6 &     0.4 &  0.4 & 0.6 &     0.6 &     0.6 \\
    \bottomrule
    \end{tabular}
    \caption{\textbf{Experiment 1:} Distribution parameters for category 1 ($\mu$) and 2 ($\upsilon$) for each task in $\mathcal{C}_S$ and $\mathcal{C}_P$. Subscript 0 is spatial frequency; subscript 1 is orientation.}\label{table:manual}
\end{table*}

\begin{table*}
    \centering
    \begin{tabular}{lrrrrrrrr}
    \toprule
    {} & \multicolumn{4}{l}{Task 1} & \multicolumn{4}{l}{Task 2} \\
    Curriculum &   $\mu_0$ &  $\mu_1$ & $\upsilon_0$ & $\upsilon_1$ &   $\mu_0$ &  $\mu_1$ & $\upsilon_0$ & $\upsilon_1$ \\
    \midrule
    0  &  0.321 & 0.339 &     0.521 &     0.339 &  0.560 & 0.267 &     0.760 &     0.267 \\
    1  &  0.236 & 0.614 &     0.436 &     0.614 &  0.410 & 0.435 &     0.610 &     0.435 \\
    2  &  0.303 & 0.768 &     0.503 &     0.768 &  0.249 & 0.524 &     0.449 &     0.524 \\
    3  &  0.363 & 0.204 &     0.563 &     0.204 &  0.469 & 0.431 &     0.669 &     0.431 \\
    4  &  0.246 & 0.615 &     0.446 &     0.615 &  0.477 & 0.520 &     0.677 &     0.520 \\
    5  &  0.558 & 0.642 &     0.758 &     0.642 &  0.346 & 0.509 &     0.546 &     0.509 \\
    6  &  0.278 & 0.743 &     0.478 &     0.743 &  0.461 & 0.572 &     0.661 &     0.572 \\
    7  &  0.363 & 0.238 &     0.563 &     0.238 &  0.315 & 0.483 &     0.515 &     0.483 \\
    8  &  0.598 & 0.556 &     0.798 &     0.556 &  0.411 & 0.722 &     0.611 &     0.722 \\
    9  &  0.439 & 0.408 &     0.639 &     0.408 &  0.268 & 0.591 &     0.468 &     0.591 \\
    10 &  0.387 & 0.478 &     0.587 &     0.478 &  0.456 & 0.238 &     0.656 &     0.238 \\
    11 &  0.506 & 0.301 &     0.706 &     0.301 &  0.256 & 0.300 &     0.456 &     0.300 \\
    12 &  0.212 & 0.439 &     0.412 &     0.439 &  0.306 & 0.671 &     0.506 &     0.671 \\
    13 &  0.328 & 0.330 &     0.528 &     0.330 &  0.521 & 0.489 &     0.721 &     0.489 \\
    14 &  0.386 & 0.661 &     0.586 &     0.661 &  0.319 & 0.420 &     0.519 &     0.420 \\
    15 &  0.446 & 0.510 &     0.646 &     0.510 &  0.340 & 0.283 &     0.540 &     0.283 \\
    16 &  0.299 & 0.544 &     0.499 &     0.544 &  0.545 & 0.502 &     0.745 &     0.502 \\
    17 &  0.525 & 0.456 &     0.725 &     0.456 &  0.444 & 0.219 &     0.644 &     0.219 \\
    18 &  0.379 & 0.777 &     0.579 &     0.777 &  0.584 & 0.634 &     0.784 &     0.634 \\
    19 &  0.432 & 0.240 &     0.632 &     0.240 &  0.226 & 0.381 &     0.426 &     0.381 \\
    20 &  0.309 & 0.395 &     0.509 &     0.395 &  0.217 & 0.627 &     0.417 &     0.627 \\
    21 &  0.237 & 0.725 &     0.437 &     0.725 &  0.448 & 0.591 &     0.648 &     0.591 \\
    22 &  0.545 & 0.269 &     0.745 &     0.269 &  0.351 & 0.426 &     0.551 &     0.426 \\
    23 &  0.211 & 0.529 &     0.411 &     0.529 &  0.408 & 0.683 &     0.608 &     0.683 \\
    24 &  0.405 & 0.452 &     0.605 &     0.452 &  0.436 & 0.700 &     0.636 &     0.700 \\
    25 &  0.443 & 0.675 &     0.643 &     0.675 &  0.229 & 0.547 &     0.429 &     0.547 \\
    26 &  0.273 & 0.538 &     0.473 &     0.538 &  0.522 & 0.525 &     0.722 &     0.525 \\
    27 &  0.339 & 0.510 &     0.539 &     0.510 &  0.353 & 0.260 &     0.553 &     0.260 \\
    28 &  0.473 & 0.687 &     0.673 &     0.687 &  0.466 & 0.437 &     0.666 &     0.437 \\
    29 &  0.475 & 0.694 &     0.675 &     0.694 &  0.536 & 0.451 &     0.736 &     0.451 \\
    \bottomrule
    \end{tabular}
    \caption{\textbf{Experiments 2 and 3:} Distribution parameters for category 1 ($\mu$) and 2 ($\upsilon$) for both tasks in each randomized grating classification curriculum. Subscript 0 is spatial frequency; subscript 1 is orientation.}\label{table:randomised}
\end{table*}

\begin{table*}
    \parbox{.4\textwidth}{
    \centering
    \begin{tabular}{lrrrr}
    \toprule
    {} & \multicolumn{2}{l}{Task 1} & \multicolumn{2}{l}{Task 2} \\
    {} &    $y_0$ & $y_1$ &    $y_0$ & $y_1$ \\
    Curriculum &        &     &        &     \\
    \midrule
    0          &      3 &   7 &      6 &   1 \\
    1          &      3 &   5 &      2 &   1 \\
    2          &      2 &   6 &      4 &   0 \\
    3          &      3 &   0 &      1 &   6 \\
    4          &      4 &   6 &      2 &   1 \\
    5          &      8 &   5 &      0 &   3 \\
    6          &      2 &   6 &      0 &   4 \\
    7          &      7 &   0 &      4 &   2 \\
    8          &      2 &   6 &      7 &   4 \\
    9          &      3 &   5 &      9 &   1 \\
    10         &      8 &   0 &      7 &   4 \\
    11         &      7 &   8 &      9 &   0 \\
    12         &      9 &   3 &      6 &   4 \\
    13         &      0 &   5 &      9 &   8 \\
    14         &      1 &   3 &      7 &   0 \\
    15         &      5 &   9 &      2 &   3 \\
    16         &      8 &   5 &      4 &   3 \\
    17         &      5 &   7 &      3 &   3 \\
    18         &      2 &   1 &      3 &   0 \\
    19         &      4 &   0 &      8 &   6 \\
    20         &      3 &   2 &      9 &   1 \\
    21         &      3 &   7 &      1 &   2 \\
    22         &      9 &   0 &      4 &   2 \\
    23         &      7 &   5 &      9 &   4 \\
    24         &      5 &   5 &      9 &   1 \\
    25         &      3 &   7 &      5 &   0 \\
    26         &      4 &   0 &      8 &   9 \\
    27         &      5 &   0 &      4 &   7 \\
    28         &      8 &   4 &      6 &   5 \\
    29         &      0 &   8 &      4 &   2 \\
    \bottomrule
    \end{tabular}
    \caption{\textbf{Experiment 4:} Selected digit pairs ($y_0$, $y_1$) for both tasks in each randomized split MNIST classification curriculum.}\label{table:mnist}
    }
    \hfill
    \parbox{.4\textwidth}{
        \centering

    \begin{tabular}{lrrrr}
    \toprule
    {} & \multicolumn{2}{l}{Task 1} & \multicolumn{2}{l}{Task 2} \\
    {} &    $y_0$ & $y_1$ &    $y_0$ & $y_1$ \\
    Curriculum &        &     &        &     \\
    \midrule
    0          &      8 &   4 &      5 &   2 \\
    1          &      8 &   3 &      5 &   2 \\
    2          &      4 &   1 &      2 &   7 \\
    3          &      4 &   2 &      6 &   9 \\
    4          &      8 &   1 &      0 &   9 \\
    5          &      6 &   4 &      1 &   8 \\
    6          &      7 &   4 &      5 &   3 \\
    7          &      7 &   4 &      9 &   6 \\
    8          &      1 &   7 &      3 &   5 \\
    9          &      3 &   5 &      9 &   6 \\
    10         &      5 &   1 &      2 &   4 \\
    11         &      1 &   2 &      5 &   8 \\
    12         &      5 &   1 &      7 &   0 \\
    13         &      5 &   9 &      2 &   4 \\
    14         &      5 &   2 &      8 &   3 \\
    15         &      6 &   9 &      2 &   3 \\
    16         &      9 &   0 &      7 &   8 \\
    17         &      3 &   1 &      2 &   4 \\
    18         &      0 &   4 &      9 &   5 \\
    19         &      8 &   5 &      9 &   4 \\
    20         &      0 &   7 &      1 &   5 \\
    21         &      4 &   2 &      6 &   9 \\
    22         &      9 &   7 &      3 &   4 \\
    23         &      8 &   6 &      2 &   0 \\
    24         &      9 &   8 &      2 &   0 \\
    25         &      6 &   0 &      3 &   8 \\
    26         &      0 &   6 &      5 &   4 \\
    27         &      2 &   4 &      7 &   5 \\
    28         &      3 &   9 &      4 &   5 \\
    29         &      8 &   2 &      4 &   7 \\
    \bottomrule
    \end{tabular}
    \caption{\textbf{Experiment 5:} Selected label pairs ($y_0$, $y_1$) for both tasks in each randomized split CIFAR10 classification curriculum.}\label{table:cifar}
    }
\end{table*}

\clearpage

\bibliographystyle{named}
\bibliography{bibliography}


\end{document}